\documentclass[sigconf]{acmart}
\AtBeginDocument{%
  \providecommand\BibTeX{{%
    \normalfont B\kern-0.5em{\scshape i\kern-0.25em b}\kern-0.8em\TeX}}}

\usepackage{enumitem}   
\usepackage{stmaryrd}
\usepackage{algorithm}
\usepackage{algorithmic}
\usepackage{dsfont}
\usepackage{url} 
\usepackage{wrapfig}
\usepackage{graphicx}
\usepackage{textcomp}
\usepackage{xcolor}
\usepackage{mathtools}
\usepackage{threeparttable}
\usepackage{autonum}
\usepackage{tabularx}
\newcolumntype{R}{>{\raggedleft\arraybackslash}X}
\usepackage{enumitem}
\usepackage{microtype}
\usepackage{subcaption}
\usepackage{booktabs} 

\usepackage{tabularx}
\usepackage{caption}
\usepackage{amsmath}

\usepackage{bbm}
\usepackage{bm}

\begin{document}

\title{Walk-and-Relate: A Random-Walk-based Algorithm for Representation Learning on Sparse Knowledge Graphs}

\author{Saurav Manchanda}
\email{manch043@umn.edu}

\renewcommand{\shortauthors}{Saurav Manchanda}

\begin{abstract}
  Knowledge graph (KG) embedding techniques use structured relationships between entities to learn low-dimensional representations of entities and relations. The traditional KG embedding techniques (such as TransE and DistMult) estimate these embeddings via simple models developed over observed KG triplets. These approaches differ in their triplet scoring loss functions. As these models only use the observed triplets to estimate the embeddings, they are prone to suffer through data sparsity that usually occurs in the real-world knowledge graphs, i.e., the lack of enough triplets per entity. In this paper, we propose an efficient method to augment the number of triplets to address the problem of data sparsity. We use random walks to create additional triplets, such that the relations carried by these introduced triplets correspond to the metapath (composition of the sequence of underlying relations) induced by the random walks. We also provide approaches to accurately and efficiently choose the informative metapaths from the possible set of metapaths, generated by the random walks. The proposed approaches are model-agnostic, and the augmented training dataset can be used with any KG embedding approach out of the box. Experimental results obtained on the benchmark datasets show the advantages of the proposed approach.
\end{abstract}




\maketitle

\section{Introduction}
Knowledge graphs (KGs) are data structures that store information about different entities (nodes) and their relations (edges). They are used to organize information in many domains such as music~\cite{fell2022wasabi}, movies~\cite{orlandi2018leveraging}, (e-)commerce~\cite{zalmout2021all}, and sciences~\cite{ioannidis2020drkg}. A common approach of using KGs in various information retrieval and machine learning tasks is to compute knowledge graph embeddings (KGE)~\cite{wang2017knowledge,goyal2018graph}. These approaches embed a KG’s entities and relation types associated with an edge into a $d$-dimensional space such that the embedding vectors associated with the entities and the relation types satisfy a pre-determined mathematical model. Numerous models for computing knowledge graph embeddings have been developed, such as TransE~\cite{bordes2013translating}, TransR~\cite{lin2015learning} and DistMult~\cite{yang2014embedding}.

There are many possible facts and relations in the real world, but collecting them can be challenging; consequently, existing knowledge graphs lack completeness. Learning embeddings for entities and relations in knowledge graphs can be considered knowledge induction, and the induced embeddings can be used to infer new triplets in the knowledge graph. The embedding quality is heavily reliant on the quantity and quality of the underlying triplets. This makes obtaining meaningful embeddings from sparse KG difficult, because of the lack of enough number of triplets per entity or relation. In addition, KG embedding approaches that use negative sampling encounter increased false negative rate in the chosen negative samples when dealing with sparse KGs. Negative sampling is performed by creating synthetic observations (relating two entities using a relation) such that the created observation is not observed in the training data. The chances of selecting an actual positive observation as a negative sample is higher with a sparse KG.

One solution to address the sparsity of triplets is to perform data augmentation. Various forms of data augmentation are possible. For example, prior approaches have explored augmenting the KG triplets with text corpus~\cite{wang2016text, ho2018rule}, using schema and semantic constraints~\cite{xie2016representation, krompass2015type}, using inference rules on the KG~\cite{guo2016jointly,zhang2019iteratively,zhao2020structure,ho2018rule}, etc. The approaches proposed in this paper resemble the ones that leverage the KG inference rules (multi-hop reasoning, in particular) to augment the KG triplets. Specifically, multi-hop reasoning is the task of learning explicit inference patterns in a KG. For example, if the KG includes the facts such as b is a's mother, c is b's husband, we want to learn the following pattern: LHS: isMother(a, b) $\And$ isHusband(b, c) $\implies$ RHS: isFather(a, c). Various approaches leverage such reasoning to expand the set of KG triplets, and hence, learn higher quality embeddings. The question we answer is: \emph{what if the isFather relation (RHS in the above pattern) is not present in the KG? Could we still leverage the composite relation isMother(a, b) $\And$ isHusband(b, c) (LHS in the above pattern) to augment the triplets?} This paper proposes a solution in this direction. Our proposed approach performs data augmentation predicated on composite relations, even if the composite relations could not be mapped to existing relations.

Specifically, we provide approaches that use random walks to efficiently and accurately estimate \emph{informative} (multi-hop) composite-relations (also called metapaths) in the KG. Each metapath can be effectively considered a new relation in the KG, thus augmenting the number of triplets in the KG. However, the difficulty in considering each metapath as a separate relation lies in the exponential size of the search space: the number of possible metapaths increases as we increase the length of the metapath. To this end, we also provide various approaches that allow parameter sharing among the relations, thus alleviating the curse of dimensionality incurred because of the exponential search space. In summary, the main contributions are as follows:
\begin{itemize}[leftmargin=*]
    \item We propose a quantitative way to measure the \emph{metapath information}, i.e., the information encoded by a metapath.
    \item We introduce computationally efficient and accurate approaches to use random walks to estimate \emph{informative} metapaths in the KG and use them to augment the KG triplets.
    \item We provide various parameter sharing approaches to tackle the curse of dimensionality induced by the exponential search space of the possible metapaths.
    \item We experimentally evaluate the performance of proposed framework on different knowledge graph benchmarks. The proposed approach outperforms the baselines.
\end{itemize}
The rest of the paper is organized as follows. We discuss the related work in Section~\ref{sec:related}, followed by the background and notation in Section~\ref{sec:notation}. In Section ~\ref{sec:proposed}, we describe the details of the proposed Walk-and-Relate method. We describe our evaluation methodology in Section~\ref{sec:methodology}. Finally, we present experimental results in Section \ref{sec:results} and conclude our discussion in Section \ref{sec:conclusion}.

\section{Related work}\label{sec:related}
The research areas relevant to the work present in this paper belong to \emph{Knowledge Graph Embeddings}, \emph{Inference Rule Mining}, and \emph{Data Augmentation for KGE} approaches. We discuss these areas below:
\subsection{Knowledge Graph Embeddings}
Knowledge graph embedding (KGE) is a machine learning task of learning a low-dimensional representation of a knowledge graph's entities and relations while preserving their semantic meaning. Leveraging their embedded representation, knowledge graphs (KGs) can be used for various applications such as link prediction, triple classification, entity recognition, clustering, and relation extraction~\cite{ji2015knowledge, abu2021relational}. KGE differs from ordinary relation inference as the information in a knowledge graph is multi-relational and more complex to model and computationally expensive. 
Many models have been devoted to estimating KGE. Some translation-based embedding models, such as TransE~\cite{bordes2013translating}, TransH~\cite{wang2014knowledge}, TransR~\cite{lin2015learning} and TransD~\cite{ji2015knowledge} apply simple linear or bilinear operations to model the embeddings of entities and relations. DistMult~\cite{yang2014embedding} and ComplEx~\cite{trouillon2016complex} design similarity scoring functions to learn semantic information. ConvE~\cite{dettmers2018convolutional} and ConvKB~\cite{nguyen2017novel} apply convolutional neural network to learn non-linear features.

In addition, there are approaches that tackle representation learning on homogeneous graphs, i.e., graphs with just one type of relation. Examples include DeepWalk~\cite{perozzi2014deepwalk} and node2vec~\cite{grover2016node2vec}. These approaches are based on random walks for embedding generation. For each node in the graph, these models generate a random path of connected nodes. The random path gives a sequence of nodes, and we can train models like Word2Vec (skip-gram)~\cite{mikolov2013distributed} to obtain the node embeddings. The approaches presented in this paper are motivated from these random-walk based embedding generation approaches.

\subsection{Rule mining}
Rule mining in KGs finds rules such as isMarried(a,b) $\implies$ staysTogether(a,b). Approaches like Inductive Logic Programming (ILP) and AMIE~\cite{galarraga2013amie} are popular choices for learning such rules. AMIE is an improved version of a fuzzy Horn rule-mining algorithm based on inductive logic programming (ILP). It overcomes the difficulty of traditional rule-mining algorithms in generating negative evidence on the basis of the open-world assumption. However, the difficulty in finding such rules lies in the exponential size of the search space: every relation can potentially be combined with every other relation in a rule. To tackle this, several extensions of the AMIE have been proposed~\cite{galarraga2015fast, lajus2020fast}.

One particular type of rule mining finds its use in the multi-hop reasoning.
Multi-hop reasoning is the task of learning explicit inference patterns in a KG. For example, if the KG includes the facts such as b is a's mother, c is b's husband, we want to learn the following rule (pattern): isMother(a, b) $\And$ isHusband(b, c) $\implies$ isFather(a, c).  Path-Ranking Algorithm (PRA)~\cite{lao2011random} is a popular algorithm for estimating such patterns in
KGs. \citet{lao2011random} uses a soft inference procedure based on a combination of constrained, weighted, random walks through the knowledge base graph to infer rules for the KG. Their solution can learn to infer different target relations by tuning the weights associated with random walks that follow different paths through the graph, using a version of the PRA.\citet{wang2015rdf2rules} proposes a rule learning approach RDF2Rules for RDF KGs. Rules are learned by finding frequent predicate cycles in RDF graphs. A new confidence measure is also proposed for evaluating the reliability of the mined rules.\citet{barati2016swarm} intro an approach called SWARM (Semantic Web Association Rule Mining) to mine semantic association rules from RDF KG. SWARM leverages the schema information to enrich the semantics of the rules. \cite{xiong2017deeppath} proposes a reinforcement learning framework to find reasoning paths in the KG. Unlike random walk-based models, the RL model allows to control the properties of the found paths. These effective paths can also be used as an alternative to PRA in many path-based reasoning methods.
\subsection{Data augmentation for KGE}
Various approaches have been developed to augment the KG data. We summarize such approaches in this section. \citet{wang2016text} augments the KG data with a text corpus. Specifically, they introduce additional text-based information and performed joint model training by combining triplets from a knowledge graph with entity descriptions in text. \citet{guo2016jointly} proposed a method that embeds rules and triplets jointly. This method learns rules based on the TransE model and retrains the embeddings combined with rules. \citet{zhang2019iteratively} proposed IterE, to alleviate the sparsity of entities in KG. In this framework, embeddings and rules are learned iteratively; the embeddings are learned from existing triplets, and new triplets are inferred from the rules that have been learned. \citet{zhao2020structure} iteratively trains embeddings and discovers patterns between relations to construct new triplets. For example, the model can discover  the pattern that $located\_in$ is the inverse relation of $has\_city$. If the KG contains the triplet $(seattle,located\_in,usa)$, then according to the discovered rules, the model can infer another triplet, $(usa, has\_city, seattle)$.~\citet{ho2018rule} iteratively extends rules induced from a KG by relying on feedback from a precomputed embedding model over the KG and external information sources including text corpora.\citet{xie2016representation} adds entity-level information into the representation learning model by using an encoder. This approach can be used to create fused semantic vectors for entities at different levels to improve the representation effect. \citet{krompass2015type} adds prior knowledge about the semantics of relation-types, extracted from the schema of the knowledge graph (type-constraints) or approximated through a local closed-world assumption, to the representation learning model. Only triplets that satisfy the type constraints are considered for modelling; thus, irrelevant or erroneous samples can be effectively filtered out to a certain extent. 
\section{Definitions and Notations}\label{sec:notation}

\begin{table}[!t]
\small
\centering
  \caption{Notation used throughout the paper.}
  \begin{tabularx}{\columnwidth}{lX}
    \hline
Symbol   & Description \\ \hline
$\mathcal{G}$    & A (knowledge) graph. \\
$\mathcal{V}$    & Set of nodes or entities in the graph $\mathcal{G}$. \\
$\mathcal{E}$    & Set of edges in the graph $\mathcal{G}$. \\
$\mathcal{A}$    & Set of node types in the graph $\mathcal{G}$. \\
$\mathcal{R}$    & Set of edge types in the graph $\mathcal{G}$. \\
$\mathfrak{m}$    & A metapath in the graph $\mathcal{G}$, having the edge-type sequence $[\mathfrak{m}_1, \mathfrak{m}_2, \cdots , \mathfrak{m}_{|\mathfrak{m}|}]$, where ${|\mathfrak{m}|}$ is the length of the metapath. \\
$\tau(v)$    & Type mapping function $\tau(v) : \mathcal{V} \rightarrow \mathcal{A}$, that maps a node to its corresponding type in the graph $\mathcal{G}$. \\
$\phi(e)$    & Type mapping function $\tau(e) : \mathcal{E} \rightarrow \mathcal{R}$, that maps an edge to its corresponding type in the graph $\mathcal{G}$. \\
$n_{e}^{\mathfrak{m}} (\mathcal{G})$    & Number of distinct metapaths of type $\mathfrak{m}$ that the edge $e$ is a part of in the graph $\mathcal{G}$.  \\
$assc_{\mathfrak{m}_i\shortrightarrow\mathfrak{m}}$    & Association measure, defined as the fraction of edges of the type $\mathfrak{m}_i$ that are a part of the metapath instance for the metapath $\mathfrak{m}$. \\
$z_{\mathfrak{m}}$    & Metapath information for the metapath $\mathfrak{m}$. \\
$conf_{\mathfrak{m} \shortrightarrow q}$    & Rule-confidence for the rule $\mathfrak{m} \shortrightarrow q$, defined as the fraction of node pairs $(h, t)$ such that the triplet $(h, q, t)$ exists given that the path $(h, \mathfrak{m}, t)$ exists. \\
$r_v$    & Representation (embedding) of $v$ where $v$ is an entity, relation or basis vector.\\
\hline
\end{tabularx}
  \label{tab:notation}
\end{table}
A graph is composed of nodes and edges $\mathcal{G}(\mathcal{V}, \mathcal{E})$, where $\mathcal{V}$ is the set of nodes and $\mathcal{E}$ is the set of edges. A knowledge graph (KG) is a special type of graph whose nodes and edges have types. 
A node in a knowledge graph represents an entity and an edge represents a relation between two entities. The edges are in the form of triplets $(h,r,t)$, which indicates that a pair of entities $h$ (head) and $t$ (tail) are connected to each other via a relation (edge-type) $r$. Each node $v \in \mathcal{V}$ and each edge $e \in \mathcal{E}$ are associated with their type mapping functions $\tau(v) : \mathcal{V} \rightarrow \mathcal{A}$ and $\phi(e) : \mathcal{E} \rightarrow \mathcal{R}$, respectively. A metapath $\mathfrak{m}$ on the knowledge graph $\mathcal{G}$ is defined as a sequence of relations (edge-types)  $\mathfrak{m} = [\mathfrak{m}_1, \mathfrak{m}_2, \cdots, \mathfrak{m}_{|\mathfrak{m}|}]$, which describes a composite relation $\mathfrak{m}_1 \circ \mathfrak{m}_2\circ \cdots\circ\mathfrak{m}_{|\mathfrak{m}|}$. Given a metapath $\mathfrak{m}$, a metapath instance $p$ of $\mathfrak{m}$ is defined as a node and edge sequence in the graph following the sequence of relations defined by $\mathfrak{m}$. The notation used in this paper is summarized in the Table~\ref{tab:notation}.

\section{Proposed Approach: Walk-and-Relate (WAR)}\label{sec:proposed}
Walk-and-Relate (WAR) uses short random walks to construct sequences of connected nodes in the KG.
Similar to Deepwalk~\cite{perozzi2014deepwalk}, the random walk samples uniformly from the neighbors of the last vertex visited until the maximum length ($l_{max}$) is reached\footnote{In principle, there is no requirement for the random walks to be of the same length and we can have restarts in the random walks}. 
For each node in the graph, WAR performs a random walk, thus sampling a random path of nodes connected to it. For each pair of nodes in this random walk, we can effectively consider them \emph{neighbors} in the graph, associated by a metapath (composite relation). Thus, we use these pair of nodes, along with the associated metapath to construct additional triplets and augment the training dataset for KG embeddings. 
While the nodes in the augmented triplets can be directly used by the triplet scoring function of the desired KG embedding method, there are some unknowns regarding how to use the metapaths. Particularly:
\begin{itemize}[leftmargin=*]

    \item \emph{Does the composite relation described by the metapath provide any additional information?} Augmenting with \emph{non-informative} metapaths could bring undesired noise to the training triplets. The is expected to be the case as the number of possible metapaths grows exponential with the length of random-walk. To address this, we take motivation from the vast literature on \emph{association mining}, and provide approaches to efficiently mine \emph{informative} metapaths and quantify the information provided by them (details in Section~\ref{sec:metapath}).  
    
    \item \emph{How do we represent the metapath as a single relation which is a requirement for KG embedding models based on triplet scoring? } For each selected informative metapath ($\mathfrak{m}$), we check if the metapath can be confidently mapped to one of the existing relations in the KG (rule-mining: $\mathfrak{m}\implies r$). 
    If yes, we use the corresponding rule to generate new triplets for the relation $r$ (Note that this step is similar to earlier approaches). Details on rule-mining are covered in Section~\ref{sec:rules}. If the metapath cannot be confidently mapped to one of the existing relations, each metapath is effectively considered as new relation in the KG. However, the difficulty in considering each metapath as a separate relation lies in the exponential size of the search space. To address this curse of dimensionality, we provide multiple solutions based on generalization via parameter sharing (details in Section~\ref{sec:generalization}).
\end{itemize}

The next sections cover the WAR framework in detail.

\subsection{How informative is a metapath?}\label{sec:metapath}
We borrow ideas from the association rule mining literature~\cite{agrawal1993mining} to measure the information conveyed through a metapath, which we call metapath-information. First, we define a association measure, denoted by $assc_{\mathfrak{m}_i\shortrightarrow\mathfrak{m}}$, that measures how often an edge with type $\mathfrak{m}_i$ is a part of an metapath instance of type $\mathfrak{m}_i$. Similar to the confidence measure used in the association rule mining, we define the $assc_{\mathfrak{m}_i\shortrightarrow\mathfrak{m}}$ as the fraction of edges of the type $\mathfrak{m}_i$ that are a part of the metapath instance for the metapath $\mathfrak{m}$. Formally, the definition is given as
\begin{equation}\label{eq:association}
    assc_{\mathfrak{m}_i\shortrightarrow\mathfrak{m}} = \frac{\sum\limits_{e \in \mathcal{G}:\phi(e) = \mathfrak{m}_i} \mathds{1}(n_{e}^{\mathfrak{m}}(\mathcal{G}) > 0)}{\sum\limits_{e \in \mathcal{G}:\phi(e) = \mathfrak{m}_i} \mathds{1}},
\end{equation}
where $n_{e}^{\mathfrak{m}}(\mathcal{G})$ denotes the distinct metapath instances of type $\mathfrak{m}$ that the edge $e$ is a part of in the graph $\mathcal{G}$.
Next, we define the metapath-information of the metapath $\mathfrak{m}$ as the product of the association measures $assc_{\mathfrak{m}_i\shortrightarrow\mathfrak{m}}$ of each of its constituent relations $\mathfrak{m}_i : 1 \leq i \leq |\mathfrak{m}|$. Formally, we denote the metapath-information of $\mathfrak{m}$ as $z_{\mathfrak{m}}$ and calculate it as
\begin{equation}
    z_{\mathfrak{m}} = \prod_{i=1}^{|\mathfrak{m}|}assc_{\mathfrak{m}_i\shortrightarrow\mathfrak{m}}.
\end{equation}

Estimation of $z_{\mathfrak{m}}$ requires enumeration of all the metapath instances of $\mathfrak{m}$. Trivial enumerating these metapath instances would lead to a time and space complexity of the order of $O(|\mathcal{E}|^{|\mathfrak{m}|})$. The exponential complexity raises scalability concerns, and prohibits from estimating the metapath importance for graphs with large number of edges.

In this direction, we provide two heuristics to address the scalability problems. First, we provide an approach to prune the exploration space of the candidate metapaths, that leverages the downward-closure property of the metapath-information. Second, we provide an approach to estimate the metapath-information on a sampled graph.

\subsubsection{Pruning the exploration space}
Our goal is to select the informative metapaths, and not to calculate the metapath-information $z_{\mathfrak{m}}$ of all possible metapaths. Thus, we employ a pruning strategy, i.e., not compute $z_{\mathfrak{m}}$ for a metapath $\mathfrak{m}$ if we have \emph{evidence} that $z_{\mathfrak{m}}$ is less than a threshold. We leverage the downward-closure property of the metapath-information to compute this \emph{evidence}. 
\begin{lemma}
metapath-information is downward-closed (anti monotonic), i.e., if $\mathfrak{m} = [\mathfrak{m}_1, \mathfrak{m}_2, \cdots, \mathfrak{m}_{|\mathfrak{m}|}]$ and $\mathfrak{\hat{m}} = [\mathfrak{m}_1, \mathfrak{m}_2, \cdots, \mathfrak{m}_{|\mathfrak{m}|-1}]$, then $z_{\mathfrak{m}} <= z_{\mathfrak{\hat{m}}}$.

\emph{Proof:} Each metapath instance of type $\mathfrak{m}$ can be mapped to a metapath instance of type $\mathfrak{\hat{m}}$, by removing the edge of relation $\mathfrak{m}_{|\mathfrak{m}|}$. Thus, 
\begin{equation}
    n_{e}^{\mathfrak{m}_i} <= n_{e}^{\mathfrak{\hat{m}}_i}, \forall 1 \leq i \leq |\mathfrak{\hat{m}}| (= |\mathfrak{m}|-1)
\end{equation}
\begin{equation}
    \implies assc_{\mathfrak{m}_i\shortrightarrow\mathfrak{m}} <= assc_{\mathfrak{\hat{m}}_i\shortrightarrow\mathfrak{\hat{m}}}, \forall 1 \leq i \leq |\mathfrak{\hat{m}}|
\end{equation}
Next, the metapath-information $z_{\mathfrak{m}}$ is expressed as
\begin{equation}
    z_{\mathfrak{m}} = \prod_{i=1}^{|\mathfrak{m}|}assc_{\mathfrak{m}_i\shortrightarrow\mathfrak{m}} \leq \prod_{i=1}^{|\mathfrak{\hat{m}}|}assc_{\mathfrak{\hat{m}}_i\shortrightarrow\mathfrak{\hat{m}}} \times assc_{\mathfrak{m}_{|\mathfrak{m}|}\shortrightarrow\mathfrak{m}}
\end{equation}
\begin{equation}
    \implies z_{\mathfrak{m}} \leq z_{\mathfrak{\hat{m}}} \times assc_{\mathfrak{m}_{|\mathfrak{m}|}\shortrightarrow\mathfrak{m}} 
\end{equation}
As a fraction, the association $assc_{\mathfrak{m}_i\shortrightarrow\mathfrak{m}} \leq 1$. Thus,
\begin{equation}
    z_{\mathfrak{m}} \leq z_{\mathfrak{\hat{m}}}
\end{equation}
\end{lemma}
Following the above downward-closure property, we only need to expand the exploration space to consider the metapath $\mathfrak{m} = [\mathfrak{m}_1, \mathfrak{m}_2, \cdots, \mathfrak{m}_{|\mathfrak{m}|}]$ only if the metapath $\mathfrak{\hat{m}} = [\mathfrak{m}_1, \mathfrak{m}_2, \cdots, \mathfrak{m}_{|\mathfrak{m}|-1}]$ is informative ($z_{\mathfrak{\hat{m}}}$ is above a threshold).
\subsubsection{Estimation on a sampled graph}
Instead of estimation on the full-graph, we sample the edges from the graph, and estimate the metapath-information from there. However, metapath-information estimated from the sampled graph (with randomly sampled edges) is lower (thus, not representative) than the metapath-information on the full graph\footnote{In principle, the very notion of missing edges in a KG means that KG itself has sampled edges from the \emph{complete} KG, which we don't have access to. Thus, the metapath-information estimated from a given KG is a lower-bound on the "ideal" metapath-information.}. This is because the numerator in the Equation~\ref{eq:association} decays faster with sampling as compared to the denominator. To address this difference, we introduce a correction factor that allows us to estimate the metapath-information on the full KG, using the sampled KG.

Given an edge $e$ of type $\mathfrak{m}_i$, it does not belong to any metapath instance of type $\mathfrak{m}$ (i.e., $n_{e}^{\mathfrak{m}}(\mathcal{G_S}) = 0$) if one of the following conditions hold:
\begin{itemize}[leftmargin=*]
    \item $e$ does not belong to any metapath instance of type $\mathfrak{m}$ in the original unsampled graph, i.e., $n_{e}^{\mathfrak{m}}(\mathcal{G}) = 0$. The probability of this condition holding is 
    \begin{equation}
        p \times \mathds{1}(n_{e}^{\mathfrak{m}}(\mathcal{G}) = 0),
    \end{equation}
    where $p$ denotes the sampling probability. i.e., $p = P(e \in \mathcal{G_S})$
    \item $e$ belongs to one or more metapath instance of type $\mathfrak{m}$ in the original unsampled (full) graph, i.e., $n_{e}^{\mathfrak{m}}(\mathcal{G}) > 0$, but each of those metapath instances do not appear in the sampled graph. This is possible if at least one of the constituent edges of those metapath instances (apart from $e$) is removed because of sampling. The probability of this condition holding is 
    \begin{equation}
        \underbrace{p}_{e\text{ is retained}} \times \underbrace{n_{e}^{\mathfrak{m}}(\mathcal{G})^{(1-p^{|\mathfrak{m}|-1})}}_{\text{at least one edge is removed}} \times \mathds{1}(n_{e}^{\mathfrak{m}}(\mathcal{G}) > 0)
    \end{equation}
\end{itemize}

\noindent Combining the above two conditions, and aggregating over the observed number of edges of type $\mathfrak{m}_i$ in the sampled graph ($\mathcal{G_S}$) that do not belong to any of the metapath instance of type $\mathfrak{m}$, we get:
\begin{multline}\label{eq:sampled_eq_1}
\sum\limits_{e \in \mathcal{G_S}:\phi(e) = \mathfrak{m}_i}\mathds{1}(n_{e}^{\mathfrak{m}}(\mathcal{G_S}) = 0) =  p\sum\limits_{e \in \mathcal{G}:\phi(e) = \mathfrak{m}_i} \mathds{1}(n_{e}^{\mathfrak{m}}(\mathcal{G}) = 0)\\
     + p\sum\limits_{e \in \mathcal{G}:\phi(e) = \mathfrak{m}_i} (1-p^{|\mathfrak{m}|-1})^{n_{e}^{\mathfrak{m}}(\mathcal{G})}
\end{multline}

\noindent To simplify the above formulation, we make following three observations:
\begin{enumerate}[leftmargin=0.5cm, label=(\roman*)]
    \item Assuming $n_{e}^{\mathfrak{m}}(\mathcal{G})$ is uniformly distributed\footnote{The assumption gives a lower bound (following the AM-GM inequality property)} across the edges, we get
\begin{equation}
    n_{e}^{\mathfrak{m}}(\mathcal{G}) \approx \frac{n_{:}^{\mathfrak{m}}(\mathcal{G})}{\sum\limits_{e \in \mathcal{G}:\phi(e) = \mathfrak{m}_i}\mathds{1}(n_{e}^{\mathfrak{m}}(\mathcal{G}) > 0)},
\end{equation}
where $n_{:}^{\mathfrak{m}(\mathcal{G})}$ is the number of metapaths instances of type $\mathfrak{m}$ in the graph $\mathcal{G}$. Further, for a metapath instance present in the unsampled graph, the probability of it present in the sampled graph is $p^{|\mathfrak{m}|}$. Thus,  $p^{|\mathfrak{m}|} \times n_{:}^{\mathfrak{m}(\mathcal{G})} \approx n_{:}^{\mathfrak{m}(\mathcal{G_S})}$. 
    \item For any metapath instance present in the unsampled graph, the probability of it present in the sampled graph is $p^{|\mathfrak{m}|}$. Thus,  $p^{|\mathfrak{m}|} \times n_{:}^{\mathfrak{m}(\mathcal{G})} \approx n_{:}^{\mathfrak{m}(\mathcal{G_S})}$. 
    \item Being a part of an metapath instance or not is both mutually exclusive and exhaustive. Thus,  
    \begin{multline}
    \sum\limits_{e \in \mathcal{G}:\phi(e) = \mathfrak{m}_i} \mathds{1}(n_{e}^{\mathfrak{m}}(\mathcal{G}) = 0) + \sum\limits_{e \in \mathcal{G}:\phi(e) = \mathfrak{m}_i} \mathds{1}(n_{e}^{\mathfrak{m}}(\mathcal{G}) > 0) \\
    = \sum\limits_{e \in \mathcal{G}:\phi(e) = \mathfrak{m}_i} \mathds{1}
    \end{multline}
\end{enumerate}

\noindent Substituting the above observations in Equation~\ref{eq:sampled_eq_1} gives us
\begin{multline}\label{eq:sampled_eq_2}
p[\sum\limits_{e \in \mathcal{G}:\phi(e) = \mathfrak{m}_i} \mathds{1} - 
x + x (1-p^{|\mathfrak{m}|-1})^{\frac{n_{:}^{\mathfrak{m}}(\mathcal{G_S})}{p^{|\mathfrak{m}|}x}}] \\
     - \sum\limits_{e \in \mathcal{G_S}:\phi(e) = \mathfrak{m}_i}\mathds{1}(n_{e}^{\mathfrak{m}}(\mathcal{G_S}) = 0) = 0,
\end{multline}
where $x=\sum\limits_{e \in \mathcal{G}:\phi(e) = \mathfrak{m}_i}\mathds{1}(n_{e}^{\mathfrak{m}}(\mathcal{G}) > 0)$. Except $x$, all other variables are observed or hyperparameters ($p$) in the Equation~\ref{eq:sampled_eq_2}. Thus, various root-finding algorithms could be used to find the optimal value of $x$ that satisfy Equation~\ref{eq:sampled_eq_2}, such as Brent's~\cite{brent1971algorithm} and Golden-section search~\cite{kiefer1953sequential}. The solution $x$ is then substituted for the numerator in the Equation~\ref{eq:association} to get the corrected association measure $assc_{\mathfrak{m}_i\shortrightarrow\mathfrak{m}}$, and thus accurately estimate the metapath-information $z_{\mathfrak{m}}$.

Algorithm~\ref{alg:metapath_information} provides the pseudo-code for estimating the informative metapaths.
\subsection{How to map the informative metapath to relations in the graph?}\label{sec:rules}
For the augmentation triplets $(h, \mathfrak{m}, t)$ to be used by the desired KG embedding approach that uses triplet scoring functions, we represent the metapath $\mathfrak{m}$ as a single relation in the KG. To address this, we mine rules of the form $\mathfrak{m} \shortrightarrow q$, using the standard notion of confidence used in the association rule mining. Specifically, we define the rule confidence (denoted by $conf_{\mathfrak{m} \shortrightarrow q}$) for the rule $\mathfrak{m} \shortrightarrow q$ as the fraction of node pairs $(h, t)$ such that the triplet $(h, q, t)$ exists given that the path $(h, \mathfrak{m}, t)$ exists. Formally,
\begin{equation}
    conf_{\mathfrak{m} \shortrightarrow q} = \frac{\sum_{h,t \in \mathcal{V}}\mathds{1}((h, q, t)\in\mathcal{G} \land (h, \mathfrak{m}, t)\in\mathcal{G}))}{\sum_{h,t \in \mathcal{V}}\mathds{1}((h, \mathfrak{m}, t)\in\mathcal{G})}.
\end{equation}

For a metapath $\mathfrak{m}$, we denote the $rulemap(\mathfrak{m})$ as the mapping that holds the key-value pairs [$q, conf_{\mathfrak{m} \shortrightarrow q}$]. Since we are only interested in high-confidence rules, we add a minimum threshold on the $conf_{\mathfrak{m} \shortrightarrow q}$ for [$q, conf_{\mathfrak{m} \shortrightarrow q}$] to be added to $rulemap(\mathfrak{m})$ (say, $0.5$).

Next, for each informative-metapath $\mathfrak{m}$, we map it to an individual relation with the following setup:
\begin{enumerate}[leftmargin=0.5cm, label=(\roman*)]
    \item If the $rulemap(\mathfrak{m})$ is non-empty, we sample a relation $q$ out of $rulemap(\mathfrak{m})$, with the sampling probability $conf_{\mathfrak{m} \shortrightarrow q}$. The sampling is done independently for each augmentation triplet $(h, \mathfrak{m}, t)$ on course of the training.
    \item If the $rulemap(\mathfrak{m})$ is empty, we simply introduce a new relation for the metapath $\mathfrak{m}$ in the KG.
\end{enumerate}

\begin{algorithm}[!t]
 \caption{Estimating metapath-information}
 \begin{algorithmic}[1]
 \renewcommand{\algorithmicrequire}{\textbf{Input:}}
 \renewcommand{\algorithmicensure}{\textbf{Output:}}
 \REQUIRE Graph $\mathcal{G}$, maximum length of metapaths to discover $l_{max}$, metapath-information threshold $t$, sampling probability $p$.
 \ENSURE  Metapaths with associated metapath-information: result\_set
    \STATE Sample $p$ edges randomly from $\mathcal{G}$ as $\mathcal{G_S}$
    \STATE Represent $\mathcal{G_S}$ in relational table format ($T$), with three columns: source (src), relation (rel), and destination (dst).
    \STATE result\_set = \{\}, $\hat{T}$ = $T$
  \FOR {$l \in [2, l_{max}]$}
    \STATE $\hat{T}$ = $\hat{T}$ inner join $T$ on $\hat{T}$.dst = $T$.src
    \FOR {each metapath $\mathfrak{m}$, group $g$ in $\hat{T}$ group by rel columns}
        \STATE Initialize $z_{\mathfrak{m}} = 1$
        \FOR {each relation $\mathfrak{m}_i \in \mathfrak{m}$}
            \STATE Calculate the association $assc_{\mathfrak{m}_i\shortrightarrow\mathfrak{m}}$ with correction factor using Equations~\ref{eq:association} and~\ref{eq:sampled_eq_2}.
            \STATE Update $z_{\mathfrak{m}} = z_{\mathfrak{m}}*assc_{\mathfrak{m}_i\shortrightarrow\mathfrak{m}}$.
            \IF{$z_{\mathfrak{m}} < t$}
                \STATE Remove the instances corresponding to $\mathfrak{m}$ from $\hat{T}$, i.e., $\hat{T}$ = $\hat{T}- g$
                \STATE break
            \ENDIF
        \ENDFOR
        \IF{$z_{\mathfrak{m}} >= t$}
            \STATE Add $\{\mathfrak{m}:z_{\mathfrak{m}}\}$ to the result\_set.
        \ENDIF
    \ENDFOR
  \ENDFOR
  \RETURN result\_set
 \end{algorithmic}
 \label{alg:metapath_information}
 \end{algorithm}


\subsection{Generalization with parameter sharing}\label{sec:generalization}
If a metapath cannot be mapped to existing relations in the KG, we introduce a new relation describing the metapath in the KG. The increase in the number of unique relations would lead to rapid growth in number of parameters (relation embeddings).  Specifically, the number of parameters rises exponentially (in terms of number of original relations in the graph) with the lengths of the considered metapaths. In practice this can easily lead to over-fitting on rare metapaths (which counters our primary goal of addressing data-sparsity). To address this problem, we propose various approaches to allow parameter sharing among the embeddings for the metapaths, thus limiting the number of unique parameters and allowing for better generalization.
\subsubsection{Using composition properties of the underlying model} 
Models such as TransE~\cite{bordes2013translating} and RotatE~\cite{sun2019rotate} have triplet scoring functions that enable them to infer the composition patterns.
For example, given the metapath $\mathfrak{m} = [\mathfrak{m}_1, \mathfrak{m}_2, \cdots, \mathfrak{m}_{|\mathfrak{m}|}]$, its representation $r_{\mathfrak{m}}$ is computed as $r_{\mathfrak{m}} = r_{\mathfrak{m}_1} + r_{\mathfrak{m}_2} + \cdots + r_{\mathfrak{m}_{|\mathfrak{m}|}}$ under the assumptions of the TransE model. Similarly, the representation $r_{\mathfrak{m}}$ under the assumptions of the RotatE model is given as $r_{\mathfrak{m}} = r_{\mathfrak{m}_1} \circ r_{\mathfrak{m}_2} \circ \cdots \circ r_{\mathfrak{m}_{|\mathfrak{m}|}}$, where $\circ$ is the Hadmard (or element-wise product). Thus, we simply use the representations of the constituent relations to infer the representation of the new relation describing the metapath, without introducing any additional parameters. Note that this approach can only be applied to models that allow inference of the composition patterns.
\subsubsection{Using parametric sequence models to represent the metapath}
Parametric sequence models such as Recurrent Neural Network (RNN) are employed to generate the representation of the metapath by using the sequence of representations of the constituent relations. The only additional parameters that are introduced by this approach are the ones corresponding to the sequence model (RNN), which is a constant overhead. The RNN parameters are estimated using back-propagation from the triplet scoring loss function of any KGE model.
\subsubsection{Using basis decomposition}
Similar to Relational Graph Convolutional Network (R-GCN)~\cite{schlichtkrull2018modeling}, we use basis-decomposition as a way to share parameters among the representations for relations (and metapaths). With basis decomposition, the representation $r_{\mathfrak{m}}$ is defined as follows:
\begin{equation}
    r_{\mathfrak{m}} = \sum\limits_{b =1}^{B}a_{\mathfrak{m},b}r_b,
\end{equation}
i.e. we calculate the representation $r_{\mathfrak{m}}$ as a linear combination of basis transformations $r_b: 1\leq b \leq B$ with coefficients $a_{\mathfrak{m},b}: 1\leq b \leq B$ such that only the coefficients depend on $\mathfrak{m}$. This approach does introduce additional parameters that grow exponentially with the lengths of the considered metapaths, but the number of coefficients are much smaller in size as compared to the representation size. 

\subsection{Putting it all together}\label{sec:wr_summary}
\begin{algorithm}[!t]
 \caption{Estimating KG embeddings with Walk-and-Relate (WAR) framework}
 \begin{algorithmic}[1]
 \renewcommand{\algorithmicrequire}{\textbf{Input:}}
 \renewcommand{\algorithmicensure}{\textbf{Output:}}
 \REQUIRE Graph $\mathcal{G}$, maximum length of random-walk $l_{max}$, triplet-loss function $f$, parameter-sharing strategy $s$, dictionary with metapaths as key and metapath-information as value $\mathcal{D}_1$ and dictionary with metapaths as key and rulemaps as value $\mathcal{D}_2$.
 \ENSURE  Embeddings for each node and relation in $\mathcal{G}$
    \STATE Initialize mapping from metapaths to new relations $N= \{\}$
    \FOR {each minibatch $z$ of nodes in $\mathcal{G}$}
        \STATE Initialize minibatch-triplets $T= \{\}$
        \FOR {each node $n_1$ in minibatch $z$}
            \STATE Perform a random walk of length $l_{max}$ starting from $n_1: [n_1, n_2, \cdots, n_{l_{max}}]$.
            \FOR {each pair of nodes $(n_i, n_j)$ in the random walk $[n_1, n_2, \cdots, n_{l_{max}}]$ such that $j-i>=2$}
                \STATE Let the path from $n_i$ to $n_j$ describe the metapath $\mathfrak{m}$.
                \IF {$\mathfrak{m} \in \mathcal{D}_1$, i.e., $\mathfrak{m}$ is informative with metapath-information $z_{\mathfrak{m}}$}
                    \STATE Get $rulemap(\mathfrak{m})$ of $\mathfrak{m}$ from $\mathcal{D}_2$
                    \IF {$rulemap(\mathfrak{m})$ is non-empty}
                        \STATE Sample a relation $q$ out of $rulemap(\mathfrak{m})$, with the sampling probability $conf_{\mathfrak{m} \shortrightarrow q}$.
                        \STATE Add triplet $(n_i, q, n_j)$ with edge-weight $z_{\mathfrak{m}}\times conf_{\mathfrak{m} \shortrightarrow q}$ to the minibatch-triplets set $T$.
                    \ELSE 
                        \IF {$\mathfrak{m} \in N$}
                            \STATE $q_{new} = N[\mathfrak{m}]$.
                        \ELSE
                            \STATE Introduce a new relation $q_{new}$ for $\mathfrak{m}$.
                            \STATE Update $N$ with $N[\mathfrak{m}] = q_{new}$.
                        \ENDIF
                        \STATE Add triplet $(n_i, q_{new}, n_j)$ with edge-weight $z_{\mathfrak{m}}$ to the minibatch-triplets set $T$.
                    \ENDIF
                \ENDIF
            \ENDFOR
        \ENDFOR
        \STATE Get a sample of edges from the graph $\mathcal{G}$ and add to $T$.
        \STATE Optimize the given triplet loss function $f$ and update the node and relation embeddings using the triplet minibatch $T$ and parameter sharing strategy $s$.
    \ENDFOR
  \RETURN Embeddings for the nodes and relations in $\mathcal{G}$
 \end{algorithmic}
 \label{alg:wr_summary}
 \end{algorithm}
Given the set of informative metapaths and associated rulemaps, we use random walks to generate instances of those metapaths on the graph. These instances are used to create augmentation triplets in the KG and are directly taken as input by any downstream KG-embedding model based on triplet scoring loss functions. The KG-embedding model, thus, has access to a larger pool of training data, effectively leading to higher-quality KG embeddings. For the KGE model to be aware of the metapath-information and rule-confidence, we use them as the edge-weight for the KGE model (edge-weights are used to scale the loss function for the corresponding triplets). Specifically, if we introduce a new relation for the metapath $\mathfrak{m}$, we use $z_{\mathfrak{m}}$ as the edge-weight for the augmented triplets added as a result of metapath instances belonging to $\mathfrak{m}$. Otherwise, we use $z_{\mathfrak{m}}\times conf_{\mathfrak{m} \shortrightarrow q}$ as the edge-weight for the augmented triplets, where the metapath $\mathfrak{m}$ is mapped to the relation $q$ with rule-confidence $conf_{\mathfrak{m} \shortrightarrow q}$.
Algorithm~\ref{alg:wr_summary} provides the pseudo-code for our algorithm Walk-and-Relate.

\section{Experimental Methodology}\label{sec:methodology}
We use the Deep Graph Library - Knowledge Embeddings (DGLKE)~\cite{zheng2020dgl} toolkit to bootstrap the implementation of WAR framework. DGLKE is designed for learning at scale. It provides implementations and benchmarks on widely-used and popular KGE models, making it more easier for users to apply and develop on top of those methods. We have also open-sourced the code for the WAR framework on GitHub\footnote{\url{https://github.com/gurdaspuriya/walk-and-relate}}.
\subsection{Datasets}

\begin{table}[!t]
\small
\centering
  \caption{Knowledge graph datasets.}
  \begin{tabularx}{\columnwidth}{XXXX}
    \hline
Dataset   & \#Nodes    & \#Edges    & \#Relations \\ \hline
FB15K    &  14,951 & 592,213 & 1,345 \\
WN18    & 40,943 & 151,442 & 18 \\
\hline
\end{tabularx}
  \label{tab:datasets}
\end{table}

\begin{figure}%
    \centering
    \subfloat[\centering WN18]{{\includegraphics[width=4cm]{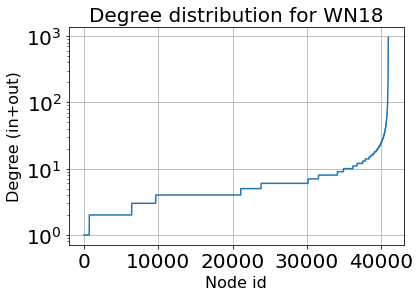} }}%
    \subfloat[\centering FB15k]{{\includegraphics[width=4cm]{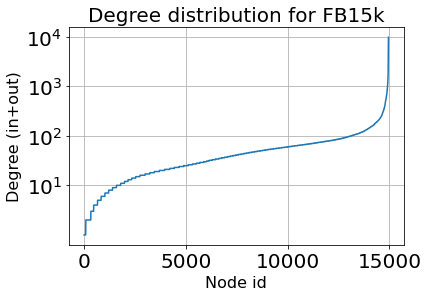} }}%
    \caption{Degree distribution of the datasets (training split)}%
    \label{fig:dataset_degree}%
\end{figure}
We used two standard benchmark datasets (WN18 and FB15K) to evaluate the performance of the proposed approaches and compare against the baselines. 
FB15k is derived from the Freebased Knowledge Graph~\cite{bollacker2008freebase}, whereas WN18 was derived from WordNet~\cite{miller1995wordnet}. 
Table~\ref{tab:datasets} shows various statistics for these datasets. Figure~\ref{fig:dataset_degree} shows the degree distribution of the two datasets. The degree distribution of FB15K is relatively more uniform (has low variance) as compared to WN18.
For WN18, $\approx60\%$ of the nodes have degree $\leq 5$, and $\approx16\%$ of the nodes have degree $\leq 2$. In contrast, FB15k has $\approx6\%$ of the nodes have degree $\leq 5$, and $\approx2\%$ of the nodes have degree $\leq 2$. Both datasets and their training/validation/test splits were used as provided by the DGL-KE toolkit.

\subsection{Metrics}
We use three standard ranking metrics~\cite{lerer2019pytorch} widely used in literature for evaluating the KG embeddings: Hit$@k$ (for $k \in \{1, 3, 10\}$), Mean Rank (MR), and Mean Reciprocal Rank (MRR). For a positive triplet $i$, let $S_i$ be the list of triplets
containing $i$ and its associated negative triplets ordered in a non-increasing score order, and let $rank_i$ be $i$th position in $S_i$. Given that, Hit$@k$ is the average number of times the positive triplet is among the $k$ highest ranked triplets; MR is the average rank of the
positive instances, whereas MRR is the average reciprocal rank of the positive instances. Formally, these metrics are defined as
\begin{equation}
    Hit@k = \frac{1}{Q}\sum\limits_{i=1}^{Q}\mathds{1}(rank_i <= k),
\end{equation}
\begin{equation}
    MR = \frac{1}{Q}\sum\limits_{i=1}^{Q}rank_i,
\end{equation}
\begin{equation}
    MRR = \frac{1}{Q}\sum\limits_{i=1}^{Q}\frac{1}{rank_i},
\end{equation}
where $Q$ is the total number of positive triplets. Note that $Hit@k$ and MRR are between
0 and 1, whereas MR ranges from 1 to the $\sum_i^Q S_i$.

\subsection{Comparison methodology}
We choose two representative KG embedding approaches based on triplet-loss functions: TransE and Distmult. TransE is a translation-based model and allows to infer compositional patterns, while Distmult is a similarity scoring-based approach. For these two approaches, we evaluate the different triplet-augmentation and parameter-sharing approaches as described below:
\begin{enumerate}
    \item No augmentation: The models are trained on explicit triplets present in the graph, and thus, no augmentation is done.
    \item WAR - Only rules: To show the promise of proposed augmentation approaches over earlier approaches that only leverage rules of the form $\mathfrak{m}\shortrightarrow q$, we provide results on an augmentation exercise where we only add additional triplets based on mapping the metapaths to the existing relations in the graph (to mimic the prior approaches). 
    \item WAR - Metapaths + No parameter sharing (WAR + MP): This approach performs augmentation with the discovered informative-metapaths, by introducing a new relation for each metapath, i.e, without any parameter sharing among the metapath (relation) embeddings.
    \item WAR - Metapaths + parameter sharing using model's composition properties (WAR + MP + Model): This approach performs augmentation with the discovered informative-metapaths, and uses the relation embeddings to compute the metapath embeddings based on the underlying composition model. Note that this approach cannot be used for DistMult as it does not allow inference of the composition patterns.
    \item WAR - Metapaths + sequence modeling for metapath (WAR + MP + RNN): This approach performs augmentation with the discovered informative-metapaths, and uses RNN as the sequence model to aggregate the representations of the constituent relations into the representation of the metapath.
    \item WAR - Metapaths + basis decomposition (WAR + MP + Basis): This approach performs augmentation with the discovered informative-metapaths, and uses the linear combination of a common bases to estimate the embeddings of the metapaths.
\end{enumerate}
\subsection{Hyperparameter selection}
We use metapaths of lengths upto three ($l_{max} = 3$) to construct augmentation triplets for the WAR approach. All approaches use the mini-batch size of $1,024$ nodes to construct random walks. For both WN18 and FB15K, we use the metapath-information threshold of $0.2$ to prune the search space. We use the full graph to estimate informative metapaths for WN18 but sample $10\%$ edges for FB15K (given its size).  To construct the rules, the rule-confidence threshold is set to $0.5$. We use random grid search to optimize for the embedding learning rate (between $[1.0, 1e-1, 1e-2, 1e-3, 1e-4, 1e-5]$), RNN learning rate (when applicable, between $[1.0, 1e-1, 1e-2, 1e-3, 1e-4, 1e-5]$), basis learning rate (when applicable, between $[1.0, 1e-1, 1e-2, 1e-3, 1e-4, 1e-5]$) and regularization norm (between $[0, 1e-10, 1e-9, 1e-8, 1e-7, 1e-6]$). We use different learning rates for embeddings, RNN parameters and basis parameters because RNN and  basis parameters get dense updates, while embeddings get sparse updates. We use MRR as the primary metric and train the KGE models until the performance on the MRR metric does not increase in two successive epochs. 
All other hyperparameters are kept the same as provided by DGL-KE for each combination of triplet loss function (TransE and Distmult) and Dataset (WN18 and FB15K).

\section{Results and Discussion}\label{sec:results}
Tables~\ref{tab:wn18_transe_results} and~\ref{tab:wn18_distmult_results} shows the performance of various approaches on the WN18 dataset, using the TransE and DistMult scoring functions, respectively. On all the metrics, the approaches that leverage data augmentation outperform the approaches without data augmentation. Particularly, we observe that levering the proposed augmentation framework (WAR) and limiting the augmentation triplets to rules (WAR - Only rules) leads to $0.32\%$ and $0.14\%$ improvement on MRR over the approach without augmentation, for the TransE and DistMult scoring functions, respectively. The improvement increases to $2.92\%$ for the TransE and $0.30\%$ for DistMult when we expand the augmentation triplets with \emph{informative} metapaths, even if those metapaths cannot be mapped to existing relation in the graph, by introducing a new relation for such metapaths. This provides evidence for our hypothesis that we can leverage the composite relations (i.e., metapaths) to provide strong learning signals to the KG embedding approaches, even though we cannot learn confident rules from these metapaths. 

Further, given that the number of distinct relations for WN18 is relatively small ($18$), we do not expect the exponential search space of metapaths to play much role in dampening the performance, thus, we do not expect much performance gain with parameter sharing approaches. This is verified from the experimental results, where we observe that leveraging the parameter sharing is either better or as good as as the situation not performing parameter sharing. With parameter sharing, performance on the WN18 for TransE tends to remain similar, but we see performance gain of up to $2.92\%$ $1.31\%$ for DistMult. 

Tables~\ref{tab:fb15k_transe_results} and~\ref{tab:fb15k_distmult_results} shows the performance of various approaches on the FB15k dataset, using the TransE and DistMult scoring functions, respectively. 
Here, we see some different trends as compared with the WN18 dataset. 
Particularly, we see that using the WAR framework and limiting the augmentation triplets to rules (WAR - Only rules) does not lead to visible performance gain over the approach without augmentation. Given that FB15k is a relatively larger dataset as compared to WN18 (3X lesser nodes and 4X more edges), the limited triplets added by rule-based augmentation does not provide enough signal as compared to the already available triplets in the graph, which explains this behavior. This is also expressed by the degree distribution plots of the two datasets in Figure~\ref{fig:dataset_degree}. WN18 has a much larger fraction of nodes with a small degree ($\approx 60\%$ of the nodes have degree $\leq 5$) as compared to FB15k ($\approx 6\%$ of the nodes have degree $\leq 5$).
We see similar trend when we use augmentation triplets with \emph{informative} metapaths as new relations as well. This can be attributed to large number of relations ($1,345$) in the FB15k dataset, which brings the curse of dimensionality given the exponential search space for metapaths. This is further evident from the performance gain with the parameter sharing strategies. Specifically, the experimental results show that leveraging the parameter sharing leads to a performance gain of $1.4350\%$ and $2.7277\%$ on the MRR metric for the TransE and DistMult scoring functions, respectively. 

\begin{figure*}%
    \centering
    \includegraphics[width=\textwidth]{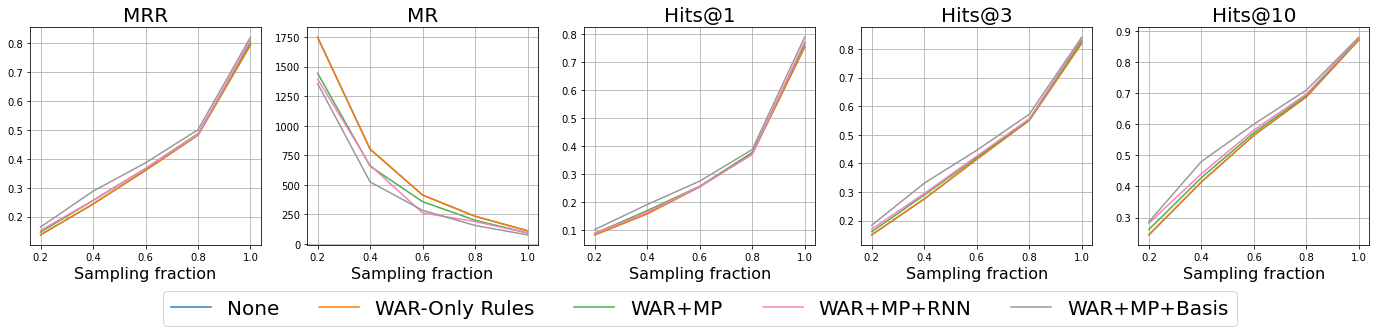} %
    \caption{Absolute performance metrics on the FB15k dataset and DistMult model with sampled training triplets.}
    \label{fig:sparsity_metrics_absolute}%
\end{figure*}

\begin{figure*}%
    \centering
    \includegraphics[width=\textwidth]{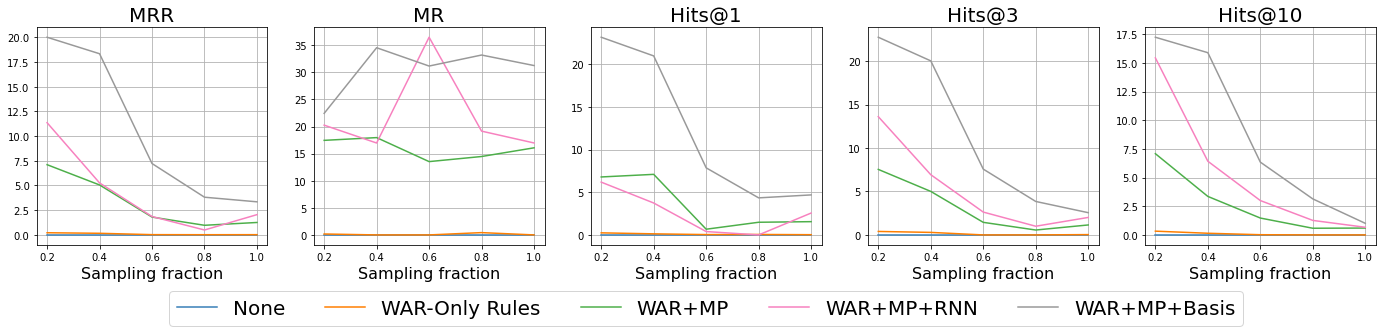} %
    \caption{Percentage performance improvement on the FB15k dataset and DistMult model with sampled training triplets.}
    \label{fig:sparsity_metrics_relative}%
\end{figure*}

Figure~\ref{fig:sparsity_metrics_absolute} shows the performance metrics for the FB15k dataset and DistMult after sampling the training dataset with different sampling probabilities: $20\%, 40\%, 60\%, 80\% and 100\%$ (original, unsampled graph). We observe that the WAR framework provides greater benefits when sampling probabilities are lower for different metrics. The WAR framework's metrics and those provided by the approach without data augmentation start converging as the sampling probability increases. To facilitate visualization, we also provide relative performance gains based on baseline approaches, i.e., approaches that do not incorporate any data augmentation. These results are provided in figure~\ref{fig:sparsity_metrics_relative}\footnote{The MR metric is unbounded, so we observe relatively high variance, and do not observe smooth patterns in the performance gain with the MR, as compared with the other considered metrics.}. For example, on the MRR metric, the WAR-framework-based approaches provide more than  $20\%$ performance gain when the training triplets are sampled with a probability of $20\%$, but the gain drops to $2.7277\%$ for the case when we do not perform sampling.
Additionally, limiting the augmentation triplets to rules (WAR - Only rules) leads to minimal gains at $20\%$, which further diminishes at higher sampling probabilities. This supports our hypothesis that we can leverage metapaths to provide strong learning signals to sparse KG embedding approaches.

The performance metrics are drastically affected by the added sparsity in the training triplets. MRR, for example, scores $\approx 0.8$ when trained on unsampled triplets, but this performance drops to $< 0.2$ when trained with $20\%$ of the training triplets. The drastic decrease is not just because of decreased training data, but because of an increased false-negative rate among the chosen negative samples. KG embedding involves negative sampling to pick out non-observed negative triplets from training data (by corrupting the entities or relations of an actual observed triplet). The chances of selecting an actual positive triplet as a negative triplet (since the positive triplet is absent from the sampled graph) increase as sparsity increases. The effect is even more pronounced with approaches that use self-adversarial negative-sampling~\cite{sun2019rotate}. By using the KGE model predictions during training, self-adversarial negative sampling selects \emph{hard} negative samples (unobserved samples that the KGE model predicts as positive). Due to the sparsity in the training triplets, the false negative rate increases, not only violating the assumptions of self-adversarial negative sampling but also causing positive triplets to be chosen as negative triplets more frequently than random triplets, which will result in poor performance.

\begin{table}[!t]
\small
\centering
  \caption{Performance on the WN18 dataset and TransE.}
  \begin{tabularx}{\columnwidth}{lRRRRR}
    \hline
Augmentation    & MRR    & MR    & HITS$@1$    & HITS$@3$    & HITS$@10$ \\ \hline
None    &  0.5886 & 211 & 0.3515 & 0.8126 & 0.9435 \\
WAR - Only rules    & 0.5905 & 211 & 0.3548 & 0.8135 & 0.9439 \\
WAR + MP    & \underline{0.6058} & 186 & 0.3841 & 0.8110 & 0.9413 \\
WAR + MP + Model    & 0.5970 & 195 & 0.3684 & 0.8082 & 0.9426 \\
WAR + MP + RNN  & \underline{0.6047} & 176 & 0.3813 & 0.8115 & 0.9406 \\
WAR + MP + Basis    & \underline{0.6041} & 190 & 0.3800 & 0.8120 & 0.9415 \\

\hline
\end{tabularx}
  \label{tab:wn18_transe_results}
\end{table}

\begin{table}[!t]
\small
\centering
  \caption{Performance on the WN18 dataset and DistMult.}
  \begin{tabularx}{\columnwidth}{lRRRRR}
    \hline
Augmentation    & MRR    & MR    & HITS$@1$    & HITS$@3$    & HITS$@10$ \\ \hline
None    &  0.8290 & 655 & 0.7351 & 0.9204 & 0.9451 \\
WAR - Only rules    & 0.8302 & 665 & 0.7372 & 0.9197 & 0.9457 \\
WAR + MP    & 0.8315 & 555 & 0.7385 & 0.9205 & 0.9478 \\
WAR + MP + RNN  & \underline{0.8399} & 571 & 0.7611 & 0.9151 & 0.9407 \\
WAR + MP + Basis    & \underline{0.8353} & 592 & 0.7446 & 0.9224 & 0.9479 \\
\hline
\end{tabularx}
  \label{tab:wn18_distmult_results}
\end{table}

\begin{table}[!t]
\small
\centering
  \caption{Performance on the FB15k dataset and TransE.}
  \begin{tabularx}{\columnwidth}{lRRRRR}
    \hline
Augmentation    & MRR    & MR    & HITS$@1$    & HITS$@3$    & HITS$@10$ \\ \hline
None    &  0.6899 & 37 & 0.5810 & 0.7734 & 0.8603 \\
WAR - Only rules    & 0.6899 & 37 & 0.5810 & 0.7737 & 0.8603 \\
WAR + MP    & 0.6843 & 39 & 0.5746 & 0.7680 & 0.8585 \\
WAR + MP + Model    & 0.6937 & 37 & 0.5803 & 0.7820 & 0.8735 \\
WAR + MP + RNN    & \underline{0.6998} & 36 & 0.5881 & 0.7869 & 0.8766 \\
WAR + MP + Basis    & \underline{0.6967} & 37 & 0.5847 & 0.7836 & 00.8753 \\
\hline 
\end{tabularx}
  \label{tab:fb15k_transe_results}
\end{table}

\begin{table}[!t]
\small
\centering
  \caption{Performance on the FB15k dataset and DistMult.}
  \begin{tabularx}{\columnwidth}{lRRRRR}
    \hline
Augmentation    & MRR    & MR    & HITS$@1$    & HITS$@3$    & HITS$@10$ \\ \hline
None    &  0.7992 & 117 & 0.7595 & 0.8240 & 0.8703 \\
WAR - Only rules    & 0.7974 & 96 & 0.7546 & 0.8237 & 0.8748 \\
WAR + MP   & \underline{0.8053} & 94 & 0.7648 & 0.8309 & 0.8761 \\
WAR + MP + RNN   & \underline{0.8107} & 93 & 0.7717 & 0.8368 & 0.8765 \\
WAR + MP + Basis    & \underline{0.8210} & 77 & 0.7876 & 0.8413 & 0.8795 \\
\hline
\end{tabularx}
  \label{tab:fb15k_distmult_results}
\end{table}

\section{Conclusion}\label{sec:conclusion}
In this paper, we address the challenge of data sparsity that usually occurs in real world knowledge graphs, i.e., the lack of enough triplets per entity. Specifically, we propose an efficient data augmentation approach, named Walk-and-Relate (WAR), to augment the number of triplets. WAR leverages random walks to create additional triplets, such that the relations carried by these introduced triplets entail from the metapath induced by the random walks. We also provide approaches to accurately and efficiently filter out informative metapaths from the possible set of metapaths, induced by the random walks. The proposed approaches are model-agnostic, and the augmented training dataset can be used with any KG embedding approach out of the box. Experimental results obtained on the benchmark datasets show the advantages of the proposed approach.

Our work makes a step towards going beyond inference rules to augment KGs, and envision that the proposed method will serve as a motivation to develop other solutions to address the sparsity in real world KGs.
\bibliographystyle{ACM-Reference-Format}
\bibliography{refs}










\end{document}